\documentclass[10pt,twocolumn,letterpaper]{article}

\usepackage{iccv}
\usepackage{times}
\usepackage{epsfig}
\usepackage{graphicx}
\usepackage{amsmath}
\usepackage{amssymb}
\usepackage{amsthm}
\usepackage{amsfonts}
\usepackage{verbatim}

\usepackage[pagebackref=true,breaklinks=true,letterpaper=true,colorlinks,bookmarks=false]{hyperref}

\iccvfinalcopy 

\def\iccvPaperID{550} 

\ificcvfinal\fi
\begin{document}

\title{What Makes Kevin Spacey Look Like Kevin Spacey}

\author{Supasorn Suwajanakorn ~~~~ Ira Kemelmacher-Shlizerman ~~~~ Steven M. Seitz\\
	\\
	University of Washington}

	\teaser{
	\includegraphics[width=.7\textwidth]{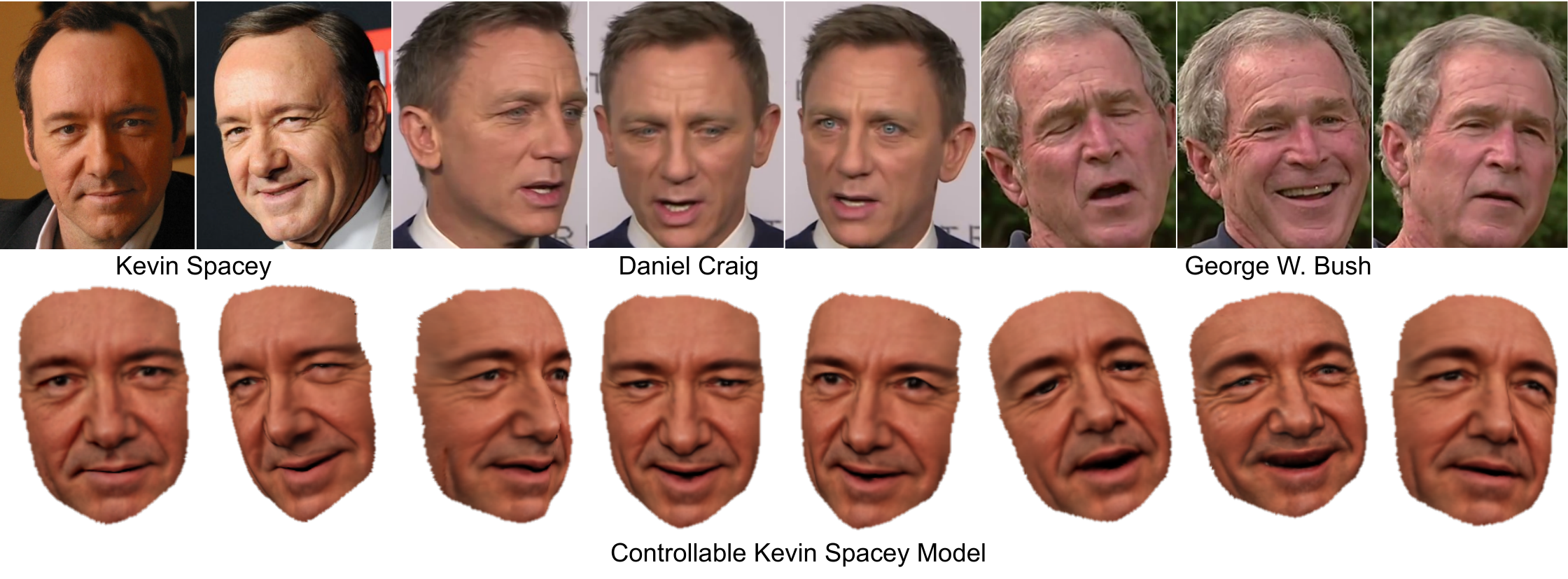} \caption{
Model of Kevin Spacey (bottom), derived from Internet Photos, is controlled by his own photos or videos of other celebrities (top).  The Kevin Spacey model captures his personality and behavior, while mimicking the pose and expression of the controllers.
}\label{fig:teaser}
}

\maketitle

\begin{abstract}
		
	We reconstruct a controllable model of a person from a large photo collection that captures his or her {\em persona}, i.e., physical appearance and behavior.  The ability to operate on unstructured photo collections enables modeling a huge number of people, including celebrities and other well photographed people without requiring them to be scanned.  Moreover, we show the ability to drive or {\em puppeteer} the captured person B using any other video of a different person A.  In this scenario, B acts out the role of person A, but retains his/her own personality and character.  Our system is based on a novel combination of 3D face reconstruction, tracking, alignment, and multi-texture modeling, applied to the puppeteering problem.  We demonstrate convincing results on a large variety of celebrities derived from Internet imagery and video.

	
\end{abstract}

\section{Introduction}
Kevin Spacey has appeared in many acting roles over the years.  He's played characters with a wide variety of temperaments and personalities.  Yet, we always recognize him as Kevin Spacey.
Why?  Is it his shape?  His appearance?  The way he moves? 

Inspired by Doersch et al's ``What Makes Paris Look Like Paris'' \cite{doer:sigg12}, who sought to capture the essence of a city, we seek to capture an actor's {\em persona}.
But what defines a persona, how can we capture it, and how will we know if we've succeeded?

Conceptually, we want to capture how a person appears in all possible scenarios.  In the case of famous actors, there's a wealth of such data available, in the form of photographic and video (film and interview) footage.  
If, by using this data, we could somehow synthesize footage of Kevin Spacey in any number and variety of {\em new roles}, and they all look just like Kevin Spacey, then we have arguably succeeded in capturing his persona. 

Rather than creating new roles from scratch, which presents all sorts of challenges unrelated to computer vision, we will assume that we have video footage of one person (actor A), and we wish to replace him with actor B, performing the same role. 
More specifically, we define the following problem:

\begin{description}
\item[\bf Input: ]  1) all available photos and/or videos of actor B, and 2) a photo collection and a single video V of actor A 
\item[\bf Output: ] a video V$'$ of actor B performing the same role as actor A in V, but with B's personality and character.
\end{description}
Figure~\ref{fig:teaser} presents example results with Kevin Spacey as actor B, and two other celebrities (Daniel Craig and George Bush) as actor A.

The problem of using one face to drive another is a form of {\em puppetry}, which has been explored in the graphics literature
e.g., \cite{sumner2004deformation, weise2009face, kholgade2011content}.  The term {\em avatar} is also used sometimes to denote this concept of a puppet.  What makes our work unique is that we derive the {\em puppet} (actor B) automatically from large photo collections.  

Our answer to the question of what makes Kevin Spacey look like Kevin Spacey is in the form of a demonstration, i.e., a system that is capable of very convincing renderings of one actor believably mimicking the behavior of another.  Making this work well is challenging, as we need to determine what aspects are preserved from actor A's performance and actor B's personality.  For example, if actor A smiles, should actor B smile in the exact same manner?  Or use actor B's own particular brand of smile?  
After a great deal of experimentation, we obtained surprisingly convincing results using the following simple recipe:  use actor B's shape,  B's texture, and A's motion (adjusted for the geometry of B's face).  Both the shape and texture model are derived from large photo collections of B, and A's motion is estimated using a 3D optical flow technique.

We emphasize that the novelty of our approach is in our application and system design, not in the individual technical ingredients.
Indeed, there is a large literature on face reconstruction \cite{debevec2012light,alexander2013digital,kemelmacher2011face,suwajanakorn2014total} and tracking and alignment techniques e.g., \cite{weise2011realtime,kemelmacher2012collection}.  Without a doubt, the quality of our results is due in large part to the strength of these underlying algorithms from prior work.  Nevertheless, the system itself, though seemingly simple in retrospect, is the result of many design iterations, and shows results that no other prior art is capable of.  Indeed, this is the first system capable of building puppets automatically from large photo collections and driving them from videos of other people.

 \begin{figure*}\center
 	\includegraphics[width=0.85\textwidth]{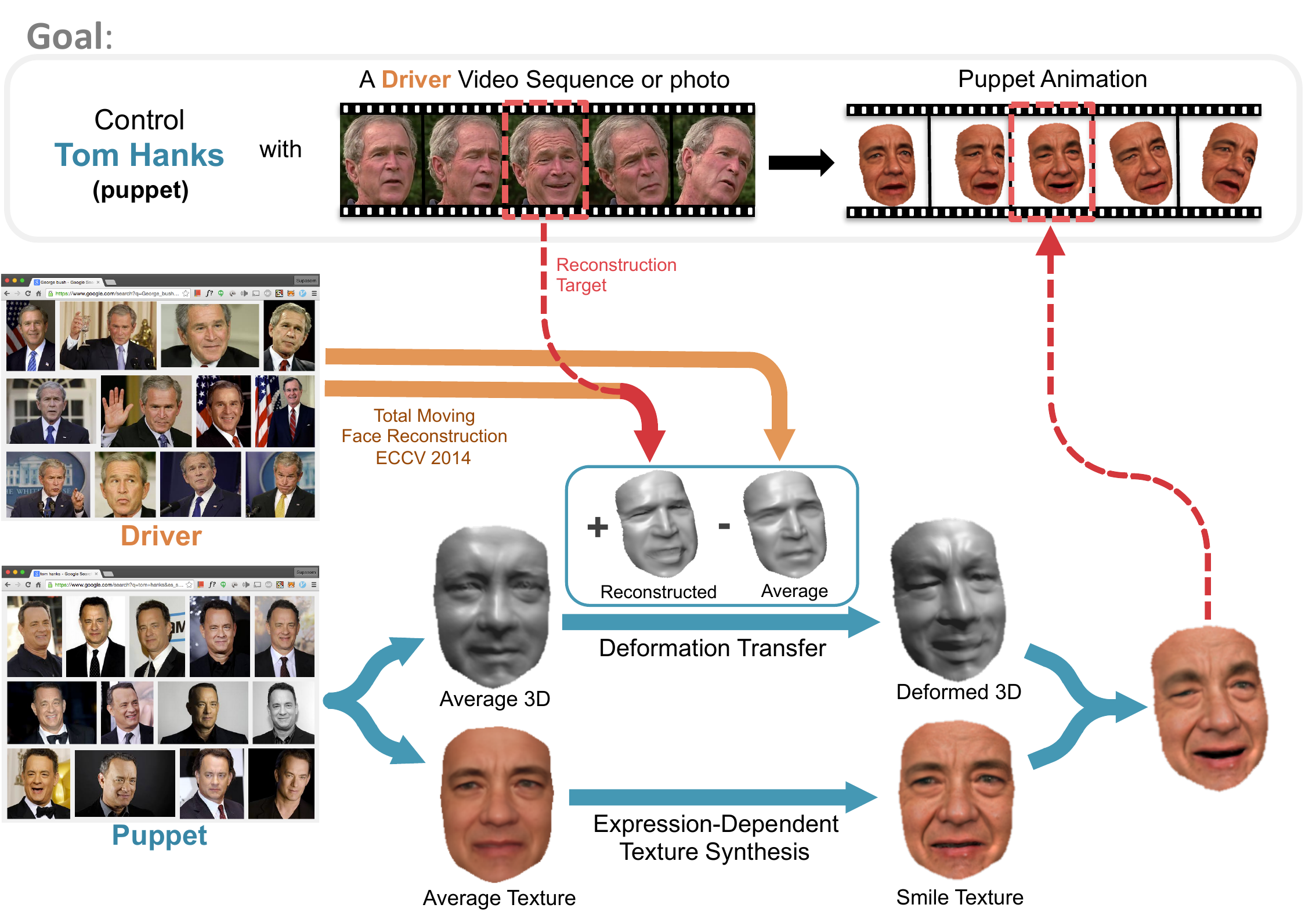}
 	\caption{One goal of our system is to create a realistic puppet of any person which can be controlled by a photo or a video sequence. Both the driver and puppet only require a 2D photo collection. To produce the final textured model, we deform the average 3D shape of the puppet reconstructed from its own photo collection to the target expression by transfering the deformation from the driver. The texture of the final model is created separately for each frame via our texture synthesize process which produces detailed, consistent, and expression-dependent textures.}\label{fig:overview}
 \end{figure*}


\section{Related Work}
Creating a realistic controllable model of a person's face is  an extremely challenging problem due to the high degree of variability in the human face shape and appearance. Moreover, the shape and texture are highly coupled:
when a person smiles, the 3D mouth and eye shape changes, and wrinkles and creases appear and disappear which changes the texture of the face. 

Most research on avatars focuses on animated, non-human faces \cite{kholgade2011content, weise2011realtime}. The canonical example is that a person drives an animated character, e.g., a dog, with his/her face. The driver’s face can be captured by a webcam or structured light device such as Kinect, the facial expressions are then transferred to a blend shape model that connects the driver and the puppet and then coefficients of the blend shape model are applied to the puppet to create a similar facial expression.  Recent techniques can operate in real-time, with a number of commercial systems now available, e.g., faceshift.com (based on \cite{weise2011realtime}), and Adobe Project Animal \cite{AdobeProjectAnimal}. 

The blend shape model that is used for non-human avatars typically capture only large scale expression deformations. Capturing fine details remains an open challenge. Some authors have explored alternatives to blend shape models for non-human characters by learning shape transfer functions  \cite{zeiler2011facial}, and dividing the shape transfer to several layers of detail \cite{xu2014controllable,kholgade2011content}.  

Creating a model of a real person, however, is an even more challenging task due to the extreme detail that is required for creation of a realistic face. One way of capturing fine details is by having the person participate in a sequence of lab sessions and use multiple synchronized and calibrated lights and camera rigs \cite{alexander2013digital}. For example, light stages were used for creation of the Benjamin Button movie--to create an avatar of Brad Pitt in an older age \cite{debevec2012light}. For this, Brad Pitt had to participate in numerous sessions of capturing every possible expression his face can make according to the Facial Action coding system \cite{ekman1977facial}. These expressions were later used to create a personalized blend shape model and transferred to an artist created sculpture of an older version of him.  This approach produces amazing results, however, requires active participation of the person and takes months to execute. 

Automatic methods for expression transfer of people used multilinear models created from 3D scans \cite{vlasic2005face} or structured light data \cite{cao2014facewarehouse}, and transfered differences in expressions of the driver's mesh to the puppet's mesh through direct deformation transfer, e.g., \cite{sumner2004deformation, weise2009face, kholgade2011content} or through coefficients that represent different face shapes \cite{weise2011realtime, vlasic2005face}, or driven by speech \cite{cao2005expressive}. These approaches either account only for large scale deformations or do not handle texture changes on the puppet.

This paper is about creating expression transfer in 3D with high detail models and accounting for texture changes that occur due to expression change. Change in texture was previously considered by \cite{liu2001expressive} via image based wrinkles transfer using ratio images, where editing of facial expression used only a single photo \cite{yang2012facial}, face swapping \cite{bitouk2008face, dale2011video}, reenactment \cite{garrido2014automatic}, and age progression \cite{kemelmacher2014illumination}. These approaches changed a person's appearance by transferring changes in texture from another person, and typically focus on a small range of expressions. Finally, \cite{kemelmacher2010being} showed that it is possible to create a puppetry effect by simply comparing two youtube videos (of the driver and puppet) and finding similarly looking (based on metrics of \cite{kemelmacher2011exploring}) pairs of photos. However, the results simply recalled the best matching frame at each time instant, and did not synthesize continuous motion.
In this paper, we show that it is possible to leverage a completely unconstrained photo collection of the person (e.g., Internet photos) in a simple but highly effective way to create texture changes, applied in 3D.

\section{Overview}

Given a photo collection of the driver and the puppet, our system (illustrated in Figure~\ref{fig:overview}) first reconstructs a rigid 3D average model  of the driver and the puppet. Next, given a video of the driver, it estimates 3D flow from each video frame to the driver's average model. This flow is then transfered onto the average model of the puppet creating a sequence of shapes that move like the driver (Section~\ref{sec:shape}). In the next stage, high detail consistent texture is generated for each frame that accounts for changes in facial expressions (Section~\ref{sec:texture}). 

%
%

\section{3D Dynamic Mesh Creation}\label{sec:shape}

\begin{figure} \center
	\includegraphics[width=0.45\textwidth]{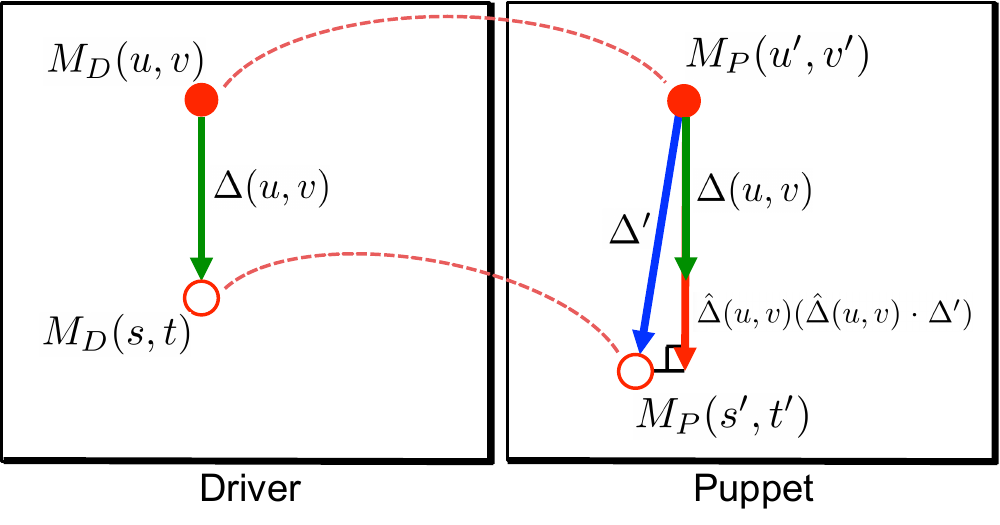}
	\caption{Magnitude adjustment in deformation transfer. Let's take an example of a blinking eye, and denote by $M_D(u,v)$  a vertex on a driver's upper eye lid. The vertex is moving down by $\Delta(u, v)$ toward $M_D(s,t)$ in order to blink. Let's denote the corresponding vertex on the puppet mesh $M_P(u',v')$. Our goal is to apply $\Delta(u, v)$ to $M_P(u',v')$, it could happen, however, that the puppet's eyes are bigger, thus we adjust the deformation and instead use $\hat{\Delta}(u, v)(\hat{\Delta}(u, v) \cdot \Delta')$.}\label{fig:flow}
\end{figure}

By searching for ``Kevin Spacey'' on Google's image search we get a large collection of photos that are captured under various poses, expressions, and lightings. In this section, we describe how we estimate an average 3D model of the driver and the puppet, and deform it according to a video or a sequence of photos of the driver. Figure~\ref{fig:flow} illustrates the shape creation process. 

\paragraph{3D Average Model Estimation.} We begin by  detection of face and fiducial points (corners of eyes, mouth, nose)  in each photo using CMU's face tracker IntraFace \cite{xiong2013supervised}.  We next align all the faces to a canonical coordinate frame  and reconstruct an average rigid 3D shape of Spacey's face. For 3D average shape reconstruction we follow \cite{kemelmacher2011face} with the modification of non-rigidly aligning photos prior to 3D reconstruction. We describe the non-rigid alignment step in Section \ref{sec:texture}. The same reconstruction pipeline is applied on the driver and the puppet photo collections, resulting in two average 3D rigid models.




\paragraph{Dynamic 3D Model.} Next, we create a dynamic model of the puppet that is deformed according to the driver's non-rigid motions. For the driver, we are given a video or sequence of photos. The first step is to reconstruct the 3D flow that deforms the driver's 3D average model to the expression of the driver in every single frame of the input video, using the method of  \cite{suwajanakorn2014total}. The geometric transformation is given as a 3D translation field $T : \mathbb{R}^3 \rightarrow \mathbb{R}^3$ applied on a driver's average shape. 

Given a reconstructed mesh at frame $i$ of a driver $M_D^i(u,v) : \mathbb{R}^2 \rightarrow \mathbb{R}^3$ parametrized on an image plane $(u, v)$ from a depth map, and the average mesh over the entire frame sequence $\overline{M}_D$, the goal is to transfer the translation field $M_D^i - \overline{M}_D$ to the puppet's base mesh $M_P$ to produce $M_P^i$. To transfer the deformation, we first establish correspondence between $M_D$ and $M_P$ through a 2D optical flow algorithm between the puppet's and driver's 2D averages from their photo collections.  

\cite{kemelmacher2012collection} has shown that we can obtain correspondence between two very different people by projecting one average onto the appearance subspace of the other by this matching illumination, and then run an optical flow algorithm between the resulting projections. With this flow, we can apply the deformation of the driver on the same facial features of the puppet. However, the direct deformation from the driver may not be suitable for the puppet, for example, if their eye sizes are different, the deformation needed to blink will be different. We solve this by scaling the magnitude of the deformation to fit each puppet as follows (Figure \ref{fig:flow}): Let the deformation vector from the driver at vertex $M_D(u,v)$, be $\Delta(u, v)$. We first find the nearest vertex to $M_D(u,v) + \Delta(u,v)$ in euclidean distance on the driver mesh, denoted by $M_D(s, t)$. Through the flow between $M_D$ and $M_P$ we computed earlier, we can establish a corresponding pair $(M_P(u',v'), M_P(s',t'))$ on the puppet mesh. The magnitude-adjusted deformation at $M_P(u', v')$ is then computed by $\hat{\Delta}(u, v)(\hat{\Delta}(u, v) \cdot \Delta')$ where $\hat{\Delta} = \frac{\Delta}{\|\Delta\|}$ and $\Delta' = M_P(s',t') - M_P(u',v')$. In addition, since the flow between the driver and puppet can be noisy around ambiguous, untextured regions, we perform the standard denoising on the term $f(u, v) = (\hat{\Delta}(u, v) \cdot \Delta')$ to obtain a regularized field $f^*(u, v)$. The particular denoising algorithm we use is ROF denoising with the Huber norm and TV regularization. The final puppet's mesh is constructed as $M_P^i(u, v) = M_P(u, v) + \hat{\Delta}(u, v)f^*(u,v)$.

\normalfont
\section{High detail Dynamic Texture Map Creation}\label{sec:texture}

In the previous section,  we have described how to create a dynamic mesh of the puppet. This section will focus on creation of a dynamic texture. The ultimate set of texture maps should be consistent over time (no flickering, or color change), have the facial details of the puppet, and change according to the driver's expression, i.e., when the driver is laughing, creases around the mouth and eye wrinkles may appear on the face. For the latter it is particularly important to account for the puppet's identity--some people may have wrinkles while others won't. Thus, a naive solution of copying the expression detail from the driver's face will generally not look realistic. Instead, we leverage a large unconstrained photo collection of the puppet's face. The key intuition is that to create a texture map of a smile, we can find many more smiles of the person in the collection. While these smiles are captured under different pose, lighting, white balance, etc. they have a common high detail that can be transfered to a new texture map.

\begin{figure*}\centering
	\includegraphics[width=0.85\textwidth]{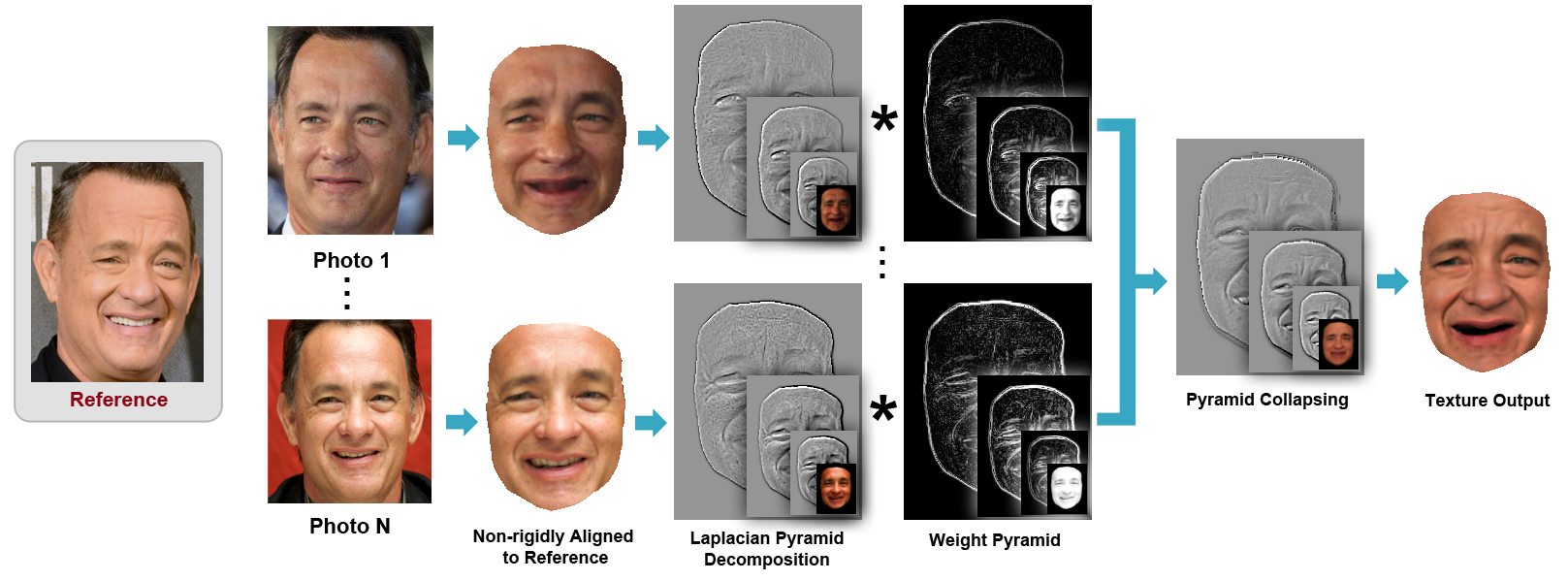}
	\caption{A diagram for synthesizing a texture for a given reference photo shown on the left. Each photo is non-rigidly aligned to the reference and decomposed into a Laplacian pyramid. The final output shown on the right is produced by computing a weighted average pyramid of all the pyramids and collapsing it. } \label{fig:pyramid}
\end{figure*}

\begin{figure*}\centering
	\includegraphics[width=1\textwidth]{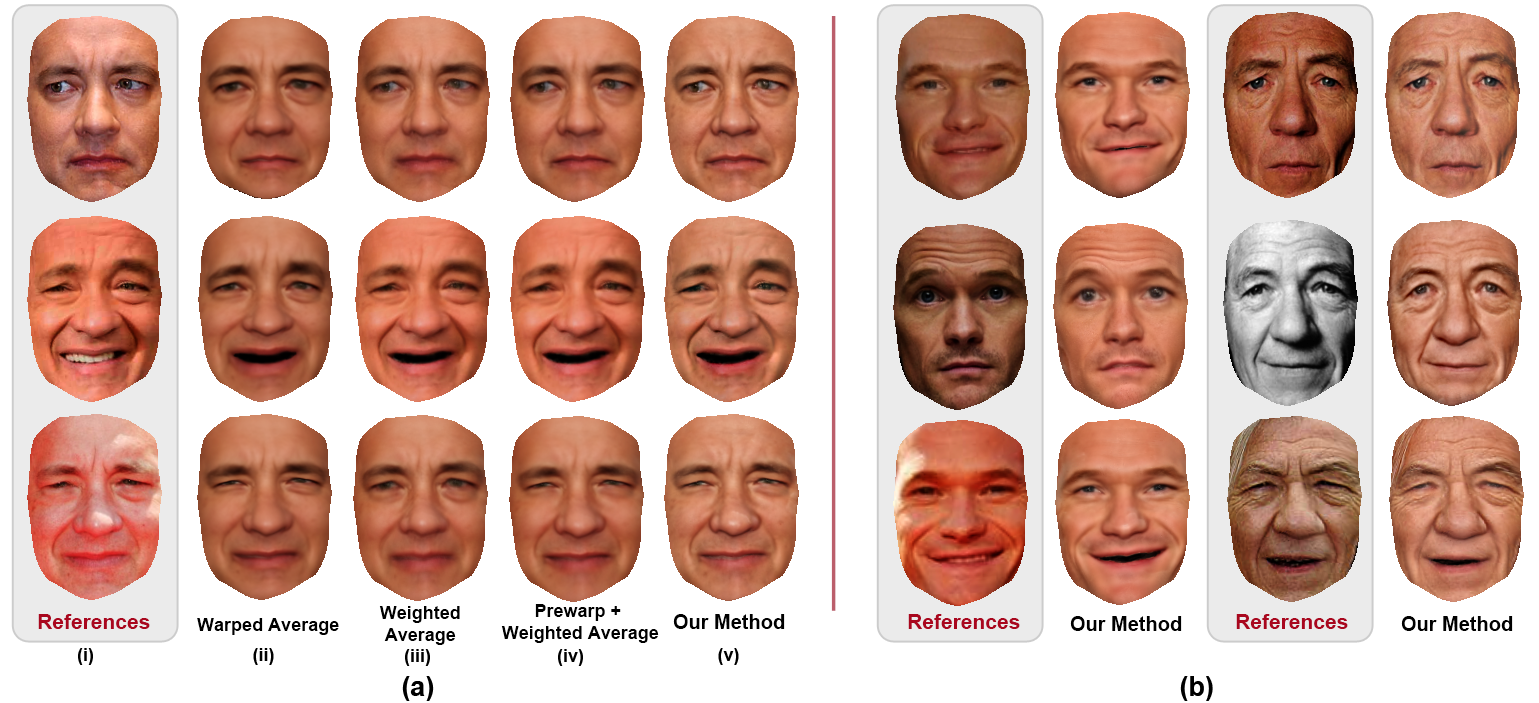}
	
	\caption{a) A comparison between our method (column v) and 3 baseline methods (columns  ii-iv) to produce a texture that matches the target expressions given in the column i. Baseline results in column (ii) are produced by warping a single average texture to the target which lack details such as creases around the mouth when the subject is smiling in the second row. Baseline results in column (iii) is produced by taking a weighed average of the photo collection with identical weights used in our method (Eq. \ref{eq:pyramid}). The facial features such as mouth appear blurry and the colors of the faces appear inconsistent. Baseline results in column (iv) are produced similarly to column (iii), but each photo is warped using thin plate spline and dense warping to the reference before taking the average. The textures appear sharper but still have inconsistent colors. Our method in column v and image b) produces consistent, sharp textures with expression-dependent details.} \label{fig:expref}
\end{figure*}

Our method works as follows. Given the target expression which is either the configuration of fiducials on the driver's face (that represents e.g., a rough shape of a smile) or by a reference photo if the driver is the same person as the puppet, we first warp all the photos in the puppet's collection to the given expression. We then create a multi-scale weighted average that perserves a uniform illumination represented by the lower frequency bands and enhances details in the higher frequency bands. Next we explain each of these in more detail. 

\paragraph{Non-rigid warping of the photos.}


\begin{figure}\centering
	\includegraphics[width=0.47\textwidth]{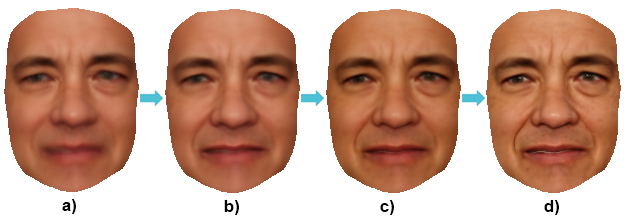}
	
	\caption{A visualization of the results after each step of the texture synthesis process to generate an average face of Tom Hanks. a) shows an average after all photos in the collection are frontalized by a 3D face template, b) after TPS warping, c) after dense warping, and d) the final texture after the multi-scale weighted average which enhances facial details. } \label{fig:averagerefine}
\end{figure}

Each photo in the puppet's photo collection has 49 fiducial points that we detected. Next we frontalize the face by marking the same fiducials on a generic 3D face model and solve a Perspective-n-Point problem to estimate the 3D pose of the face in the photo. The model is then back-projected to produce a frontal-warp version of each photo. Let the rigid pose-corrected fiducials in each photo be $F^i \in \mathbb{R}^{2\times 49}$ and the target fiducials be $F^T$. Given two sets of fiducials we estimate a smooth mapping $r$ that transforms the $i$-th photo to the target expression using a smooth variant of thin-plate splines \cite{wahba1990spline} which minimizes the following objective:
\begin{equation}
\min_r \sum_{i=1}^n \|F^i - r(F^T)\| ^2 + \lambda \iint r_{xx}^2 + 2r_{xy}^2 + r_{yy}^2 \,\textrm{d}x\,\textrm{d}y
\end{equation}
The optimal mapping $r$ satisfying this objective can be represented with a radial basis function $\phi(x) = x^2\log x$ and efficiently solved with a linear system of equations \cite{wahba1990spline}. Given the optimal $r$, we can then warp each face photo to the target expression by backward warping.  However, this warping relies only on a sparse set of fiducials and the resulting warp field can be too coarse to capture shape changes required to match the target expression which results in a blurry average around eyes and mouth (Figure \ref{fig:averagerefine} b). To refine the alignment, we perform an additional dense warping step by exploiting appearance subspaces based on \cite{suwajanakorn2014total, kemelmacher2012collection}. The idea is to warp all photos to their average, which now has the target expression, through optical flow between illumination-matched pairs. Specifically, let the $i^\text{th}$ face image after TPS warping be $I^i$, its projection onto the rank 4 appearance subspace of the TPS warped photos be $\hat{I}^i$. The refined warp field is then simply the flow from $I^i$ to $\hat{I}^i$. In the case where a reference photo is available, (Figure \ref{fig:expref}), we can further warp the entire collection to the reference photo by computing an optical flow that warps $\hat{I}^T$ to $I^T$, denoted by $F_{{\hat{I}^T} \rightarrow I^T}$, and compute the final warp field by composing $F_{{I^i} \rightarrow \hat{I}^i} \circ F_{\hat{I}^i \rightarrow {\hat{I}^T}} \circ F_{{\hat{I}^T} \rightarrow I^T} = F_{{I^i} \rightarrow \hat{I}^i} \circ F_{{\hat{I}^T} \rightarrow I^T}$ from the fact that $F_{\hat{I}^i \rightarrow {\hat{I}^T}}$ is an identity mapping.

\begin{figure}\centering
	\includegraphics[width=0.35\textwidth]{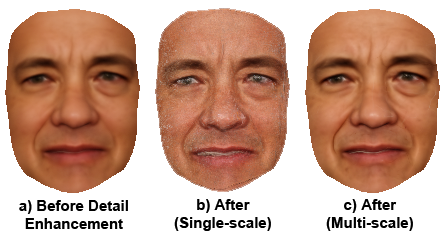}
	
	\caption{a) shows Tom Hanks' average before detail enhancement. b) and c) show the average after single-scale and multi-scale blending.  } \label{fig:multiscale}
	\vspace{-1em}
\end{figure}

\paragraph{Adding high-detail.}
Given the set of aligned face photos, we  compute a weighted average of the aligned photos where the weights measure the expression similarity to the target expression and the confidence of high-frequency details. We measure  expression similarity by $L_2$ norm of the difference between the source and target fiducial points, and  high-frequency details by the response of a Laplacian filter. A spatially-varying weight $W^i_{jk}$ for face $i$ at pixel $(j, k)$ is computed as:
\begin{equation}
W^i_{jk} = \exp\left(\frac{-\|F^T - F^i\|^2}{2\sigma^2}\right) \cdot (L^i_{jk})^\alpha
\end{equation}
where $L^i_{jk}$ is the response of a Laplacian filter on face image $i$ at pixel $(j, k)$. An average produced with this weighting scheme  produces blending artifacts, for example if high-frequency details from many photos with various illuminations are blended together (Figure \ref{fig:multiscale}). To avoid this problem, the blending is done in a multi-scale framework, which blends different image frequency separately. In particular, we construct a Laplacian pyramid for every face photo and compute the weighted average of each level from all the pyramids according to the normalized $W^i_{jk}$, then collapse the average pyramid to create a final texture. 

With real photo collections, it is rarely practical to assume that the collection spans any expression under every illumination. One problem is that the final texture for different expressions may be averaged from a subset of photos that have different mean illuminations which results in an inconsistency in the overall color or illumination of the texture. This change in the color, however, is low-frequency and is mitigated in the multi-scale framework by preferring a uniform weight distribution in the lower frequency levels of the pyramid. We achieve this is by adding a uniform distribution term, which dominates the distribution in the coarser levels:
\begin{equation}\label{eq:pyramid}
W^i_{jk} = \left(\exp\left(\frac{-\|F^T - F^i\|^2}{2\sigma^2}\right) + \tau l^{-\beta}\right) \cdot (L^i_{jk})^\alpha
\end{equation}
where $l \in \{0, \ldots, p-1\}$ and $l=0$ represents the coarsest level of a pyramid with $p$ levels, and $\tau$ and $\beta$ are constants.

\section{Experiments}
\begin{figure}\centering
	\includegraphics[width=0.47\textwidth]{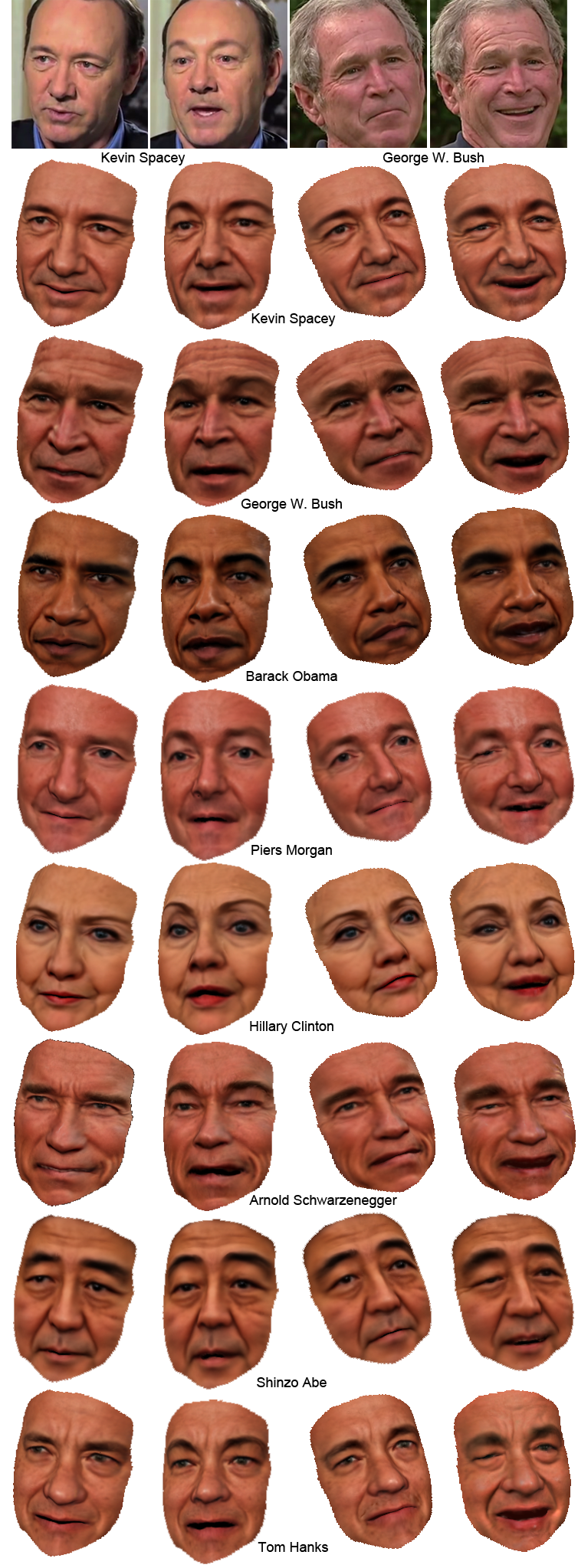}
	\caption{The first row contains two frames from YouTube videos of Kevin Spacey and George W. Bush used as referenes for puppets of many celebrities in the following rows.} \label{fig:manypeople}
\end{figure}

\begin{figure}\centering
	\includegraphics[width=0.47\textwidth]{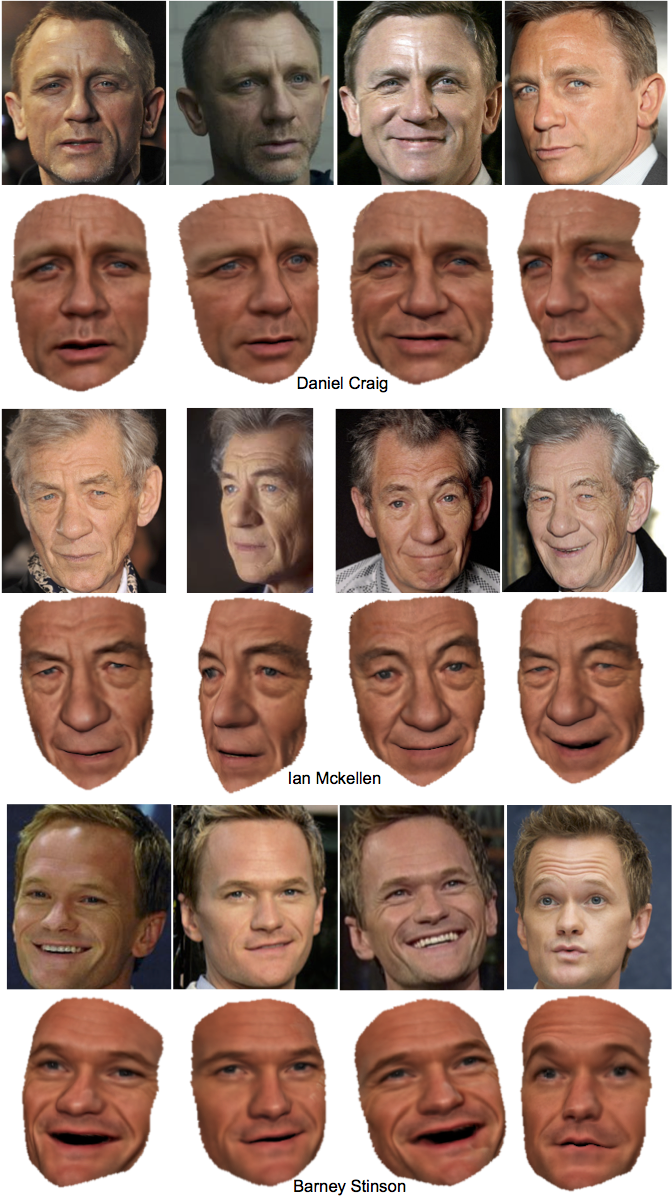}
	\caption{We show 3 example subjects for 3D shape and texture reconstruction. The input is a set of photos with varying expressions and appearances, and the output is 3D textured shapes in the same expressions as the input.} \label{fig:inputref}
	\vspace{-1.5em}
\end{figure}
In this section, we describe implementation details, runtime, and our results.

\textbf{Implementation details}
In Section \ref{sec:shape}, the 3D average models for both driver and puppet are reconstructed using \cite{suwajanakorn2014total} which outputs meshes as depth maps with a face size around 194 x 244 (width x height) pixels. To find correspondence between the driver and puppet for deformation transfer purpose, we project the 2D average of the puppet onto the rank-4 appearance subspace of the driver, then compute an optical flow using Ce Liu's implementation based on Brox et al.\cite{brox2004high} and Bruhn et al. \cite{bruhn2005lucas} with parameters ($\alpha,$ ratio, minWidth, outer-,inner-,SOR-iterations) = $(0.02, 0.85, 20, 4, 1, 40)$. The ROF denoising algorithm used for adjusting deformation magnitude has only two parameters: The weight constant for the TV regularization which is set to 1, and the Huber epsilon to 0.05. In Section \ref{sec:texture}, $\lambda$ in TPS warping objective is set to 10, and the dense warping step uses the same optical flow implementation but with $\alpha = 0.3$. For Equation $\ref{eq:pyramid}$, $(\alpha, \beta, \tau) = (1, 20, 1)$, $\sigma$ is typically set to 10 but can vary around $6-12$ for different sizes and qualities of the photo collections (See \ref{sec:discussion}).

\textbf{Runtime} We test our system on a single CPU core of a quad-core Intel i7-4770@3.40GHz. Computing a deformed puppet mesh based on the driver sequence takes 0.2 second per frame with 0.13 second spent on denoising. Synthesizing a texture which includes TPS warping, dense warping, and multi-scale pyramid blending takes 0.34 second per frame on average.

Evaluating a puppetry system objectively are extremely hard, and there exists no accuracy metric or benchmark to evalute such system. Ground-truth shapes for evaluating deformation transfer across two people cannot be captured as this requires the puppet person whose shape will be captured, to perform exactly like a driver sequence, which is not possible unless the person is the driver themselves. However, such a setup of self puppetry to evalute the reconstructed geometry requires no deformation transfer and does not evaluate our system. Evaluating the synthesized textures is also qualitative in nature as the average texture we generate cannot be pixel-wise compared to the reference. We provide results and input references for qualitative comparisons and point out areas where further improvement can be done. 

From Google Images, we gathered around 200 photos for celebrities and politicians in Figure \ref{fig:manypeople}. We generated output puppetry sequences of those people performing various facial expressions driven by YouTube Videos of Kevin Spacey and George W. Bush in the top row. These 3D models are generated by warping an average model of each person with 3D optical flow transfered from the driver (top). So, to render these texture-mapped models, we only synthesize textures in their neutral expressions for the average models but use the target expressions to calculate the blending weights. The identities of these puppets are well-preserved and remain recognizable even when driven by the same source, and the transformation provides plausible output for puppets with different genders, ethnicities, skin colors, or facial features. Facial details are enhanced and change dynamically according to the reference expressions, for example, in creases around the mouth in the last column. \textit{We strongly encourage the readers to watch our supplementary videos for these results.}

In Figure \ref{fig:expref}, we show the capability to recreate consistent textures with similar expressions as reference photos in the photo collection. In other words, we are able to ``regenerate'' each photo in the entire collection so that they appear as if the person is performing different expressions within the same video or photograph captures. Note that each reference here is part of the photo collection used in the averaging process. Texture results for references outside the photo collection is in Figure \ref{fig:inputref}. We compare our method with 3 baseline approaches: 1. A single static average is TPS warped to the reference. This approach produces textures that lack realistic changes such as wrinkles and creases, and  shapes that only roughly match the reference (e.g. eyes in column (ii) second row which appear bigger than the reference) because the warping can only rely on sparse fiducial points. 2. A weighted average of the photo collection using identical weights as our method. With this approach, creases can be seen, but the overall texture colors appear inconsistent when there is a variation in the mean color of different high-weighted sets of photos. The overall textures look blurry as there is no alignment done for each photo, and the shapes (eyes in the third row) do not match the reference when the number of similar photos in the collection is small. 3. An improved weighted average with prewarping step which includes TPS and dense warping similar to our pipeline. The prewarping step improves the shapes and the sharpness of the faces, but the textures remain inconsistent. Our method in column (v) produces sharp, realistic, and consistent textures with expression-dependent details and is able to match references with strong illuminations or in black-and-white in Figure \ref{fig:expref} (b). Since the references are part of the averaging process, some high-frequency details such as wrinkles are transfered to the output texture. However, the low-frequency details such as shading effects, soft shadow under the nose (in the last example, middle row), or highlights (in the second example, last row) are averaged out in the multi-scale blending and are not part of the final textures.

In Figure \ref{fig:inputref}, we show self-puppetry results where we render output 3D models from \cite{suwajanakorn2014total} with our textures. Similarly to Figure \ref{fig:manypeople}, we only synthesize textures in neutral expressions for the average models with blending weights calculated based on the target expressions. The reference photos are \textit{excluded} from the photo collection in the averaging process. Our textures remain consistent when the references have different lightings and look realistic from various angles. In the fourth reference in the last row, our textures have wrinkles but are less pronounced than the input reference, which is due partly to the fact that the number of photos with wrinkles in the collection is less than 5\%.

\section{Discussion}\label{sec:discussion}
The quality of the synthesized textures highy depends on many aspects of the photo collection which include the number and resolutions of the photos, expression and light varations. Since the textures are synthesized based on the assumption that we can find photos with similar expressions, the results will degrade with smaller photo collection (less expression variation). In that situation, the method needs to take into account less-similar photos with a larger standard deviation in Equation \ref{eq:pyramid} resulting in a less pronouced expression. If the standard deviation is kept small, high-frequency details can flicker when the rendered models from video input are played in sequence. Higher resolution photos directly contribute to a sharper average. Our method is less sensitive to having small light variations, in contrast to expression variations, because the shading differences are of low-frequency and can be shared across a wider range of photos in the coarser levels of pyramid. 

When a photo collection contains in the order of thousands photos such as when we extract frames from all movies starring a particular actress, additional characteristics of photos can be used to fine-tune the similarity measure in the averaging process such as the directions of lights in the scene to enable a religthing capability or the age of the person (e.g. from a regressor) to synthesize textures at different ages. Only a small modification is needed to implement these changes in our framework. It is also useful to learn the association between the apperance of facial details and facial motions to help with unseen expressions that may share common facial details with already existing photos in the collection. 

\section{Conclusion}
We presented the first system that allows reconstruction of a controllable 3D model of any person from a photo collection toward the goal of capturing persona. The reconstructed model has time-varying, expression-dependent textures and can be controlled by a video sequence of a different person. This capability opens up the ability to create puppets for any photo collection of a person, without requiring them to be scanned.  Furthermore, we believe that the insights from this approach (i.e., using actor B's shape and texture but A's motion), will help drive future research in this area.


{\small
\bibliographystyle{ieee}
\bibliography{puppetpaper_arxiv}
}

\end{document}